\documentclass[a4paper]{article}

\usepackage[english]{babel}
\usepackage[T1]{fontenc}

\usepackage[a4paper,top=3cm,bottom=2cm,left=3cm,right=3cm,marginparwidth=1.75cm]{geometry}

\usepackage{amsmath}
\usepackage{graphicx}
\usepackage{multirow}
\usepackage{authblk}
\usepackage[colorinlistoftodos]{todonotes}
\usepackage[colorlinks=true, allcolors=blue]{hyperref}

\title{An End-to-End Two-Phase Deep Learning-Based workflow to Segment Man-made Objects Around Reservoirs}
\author{Nayereh Hamidishad $^{1,}$*, and  Roberto Marcondes Cesar Junior $^{1}$}

\affil{University of São Paulo, São Paulo, SP, Brazil}

\begin{document}
\maketitle

\begin{abstract}
Reservoirs are fundamental infrastructures for the management of water resources. Constructions around them can negatively impact their quality. Such unauthorized constructions can be monitored by land cover mapping (LCM) remote sensing (RS) images. In recent years, deep learning (DL) has attracted considerable attention as a method for LCM the RS imagery and has achieved remarkable success. In this paper, we develop a new approach based on DL and image processing techniques for man-made object segmentation around the reservoirs. In order to segment man-made objects around the reservoirs in an end-to-end procedure, segmenting reservoirs and identifying the region of interest (RoI) around them are essential. In the proposed two-phase workflow, the reservoir is initially segmented using a DL model. A post-processing stage is proposed to remove errors such as floating vegetation. Next, the RoI around the reservoir (RoIaR) is identified using the proposed image processing techniques. Finally, the man-made objects in the RoIaR are segmented using a DL architecture. To illustrate the proposed approach, our task of interest is segmenting man-made objects around some of the most important reservoirs in Brazil. Therefore, we trained the proposed workflow using collected Google Earth (GE) images of eight reservoirs in Brazil over two different years. The U-Net-based and SegNet-based architectures are trained to segment the reservoirs. To segment man-made objects in the RoIaR, we trained and evaluated four possible architectures, U-Net, FPN, LinkNet, and PSPNet. Although the collected data has a high diversity (for example, they belong to different states, seasons, resolutions, etc.), we achieved good performances in both phases. The highest achieved F1-score for the test sets of phase-1 and phase-2 semantic segmentation stages are 96.53\% and 90.32\%, respectively. Furthermore, applying the proposed post-processing to the output of reservoir segmentation improves the precision in all studied reservoirs except two cases. We validated the prepared workflow with a reservoir dataset outside the training reservoirs. The F1-scores of the phase-1 semantic segmentation stage, post-processing stage, and phase-2 semantic segmentation stage are 92.54\%, 94.68\%, and 88.11\%, respectively, which show high generalization ability of the prepared workflow.\\

\textbf{keywords}: land cover mapping; deep learning; Google Earth imagery
\end{abstract}

\section{Introduction}

Reservoirs reduce the effects of interseasonal and interannual streamflow fluctuations and hence facilitate water supply, hydroelectric power generation, and flood control, to name a few \cite{gao2012global}.  
There is a significant interaction between the environment and reservoirs as essential water resource management tools. For example, reservoirs affect the quality of the water downstream of their dams, and human activities affect the quality of the reservoir's water as well as the chemical and biological processes in it \cite{votruba1989water}.

Unauthorized constructions around reservoirs can be considered destructive activities that can be monitored by LCM of RS images. The purpose of this study is to segment man-made objects around reservoirs. However, to reach this aim using an end-to-end workflow, we have to segment the reservoirs and detect the RoI around them besides segmenting the man-made objects.

The pixel-based, object-based (OB), and, recently, DL methods are three different approaches that can be implemented for LCM RS images. 
Pixel-based methods (e.g., SVM) rely on the spectral signatures of individual pixels, and each pixel is independently classified \cite{c11}. With the increase in the spatial resolution of satellite images by improving in RS systems, a single pixel does not capture well the characteristics of targeted objects, and it causes the reduction in the accuracy of classification using pixel-based methods \cite{c2}. Over the last decades, the RS community has undertaken considerable efforts to promote the use of OB technology for LCM \cite{blaschke2014geographic,khodam}. In contrast with pixel-based methods, OB classification methods are less sensitive to the spectral variance within the objects. They can use both object features and spatial relations between the objects. However, the popularity of OB methods is affected by two factors: 1- the majority of them rely on pricey commercial software; 2- The result is highly influenced by parameter selection \cite{zhang2020well}.

DL has made significant strides in recent years, enabling high-level feature extraction to be carried out automatically while displaying promising results in various domains, including image semantic segmentation. Recently, convolutional neural networks (CNNs) have been among the most advanced algorithms for the semantic segmentation of RS images, and their superior performance compared to traditional methods has been proved \cite{9376238, wurm2021deep, malerba2021continental}. The decoder-encoder networks and spatial pyramid pooling-based networks can be counted as two state-of-the-art and widely used categories of CNNs. The decoder-encoder-based networks consist of an encoder path and a decoder path. The encoder path consists of convolutional layers to extract the feature maps. Next, these features are transformed/up-sampled to dense label maps in the decoder path. U-Net, SegNet, and FPN are of this category that have demonstrated strong performances and are frequently utilized in RS semantic segmentation \cite{neupane2021deep, wang2021application}. The spatial pyramid pooling-based networks contain a pyramid pooling module to collect multi-level global information of the input image. PSPNet proposed by \cite{zhao2017pyramid} is a widely used architecture of this category \cite{zhang2020well,pspref2}. In this paper, all mentioned architectures besides LinkNet are trained and are detailed in the next section.

Semantic segmentation of water bodies is studied in several works. For example, \cite{reserp1} has fused panchromatic and RGB images and used a DL-based workflow for water body recognition. A DL encoder-decoder framework is proposed by \cite{reserp2} to extract water bodies from 4-band RS images with resolutions greater than one meter. The authors of \cite{reserp3} combine an enhanced super-pixel method with DL to extract urban water bodies from multi-spectral bands with low spatial resolutions (>4m). The RapidEye 5m resolution images are used by \cite{reserp4} to compare pixel-based methods with DL methods in segmenting gorges reservoir areas to water bodies and other land covers. The capacity of NDWI and NDSWI indices in mapping water surfaces in 4-band (near-infrared, Red, Green, and Blue) high-resolution RS images using DL and ML methods is studied by \cite{reserp5}. The utilized data belongs to National Oceanic and Atmospheric Administration (NOAA), which covers areas inside the USA.

DL is also popular among studies on semantic segmentation of man-made objects in RS images. For example, a CNN is used by \cite{manmaderef2} to classify ROSIS hyper-spectral images as man-made and non. The man-made class in this work consists of asphalt, metal sheets, bricks, bitumen, and tiles. Before feeding to the network, the data is pre-processed by Randomized Principal Component Analysis for input dimension reduction. The authors of \cite{30t2} have proposed an OB-DL framework to semantic segment two publicly available ISPRS datasets. These datasets are annotated to the impervious surface, building, low vegetation, tree, car, and clutter. Two extension versions of U-Net are proposed by \cite{builref3} to segment buildings and roads in RGB RS images with 0.5m resolution. In the proposed workflow, each class is trained in a distinct network because of the type of available ground truths. Residential land, industrial land, traffic land, woodland, and unused land are five defined classes by \cite{builref4} for collected RGB images with 0.5 m resolution. To segment images, they proposed a workflow in which the images are fed to two networks in parallel. Next, their output feature maps are fused to produce the final map. Two sites on the North Slope of Alaska are studied by \cite{builref5} using 4-band commercial satellite images with resolutions from 0.5 up to 0.87. The utilized model in this work is the U-Net with ResNet50 as the backbone.

To the best of our knowledge, the segmentation of reservoirs in RS images using DL models has not been explored in the literature. This class is always considered in a broad class termed water bodies. Furthermore, the RoIaR man-made objects segmentation has not been extensively explored. Moreover, the man-made object segmentation studies implemented on urban high-resolution RS images do not consider the countryside. Therefore, they do not contain non-asphalted roads as a challenging land cover map. 

In this study, we propose a post-processing process after segmenting reservoirs to detect errors and construct an accurate reservoir map. Furthermore, man-made objects in both urban and countryside areas are studied. We also suggest a method to detect RoIaR using image processing techniques. In the proposed two-phase workflow, we first detect the reservoirs, then the RoI, and finally segment man-made objects in the RoI. In this way, we avoid annotating and predicting areas outside the RoI. These are the main relevant contributions of this paper.

Although elevation data can improve the detection process, they are not currently viewed as a cost-effective solution to map RS images \cite{buildet}. Moreover, spatial resolution is more critical than spectral resolution in urban LCM \cite{neupane2021deep}. Therefore, we collected the data using the Google Earth platform, which is a widely used database \cite{hou2019w,zhang2020well}. In this platform, we have access to free high-resolution RS images from target reservoirs at various times. GE covers more than 25 percent of the Earth’s land surface and three-quarters of the global population by images with sub-meter resolution \cite{c1,c15}. Furthermore, the appearance of GE images is improved using color balancing, warping, and mosaic processing \cite{c11}. Therefore, it can be used for studying many other reservoirs.

The organization of our paper is as follows: Section \ref{secmethod} describes the studied reservoirs, collected data characteristics, applied data pre-processing pipeline, the proposed workflow for segmenting man-made objects around the reservoirs, and corresponding utilized methods. Next, the performance of each workflow stage, besides results visualization and workflow evaluation, is explored in section \ref{secresult}. The results and findings of the study are discussed in section \ref{secdiscuss}. Finally, the paper is concluded in section \ref{secconclud}.
\section{Materials and Methods}
\label{secmethod}
\begin{figure}[!ht]
      \centering
      \includegraphics[scale=0.42]{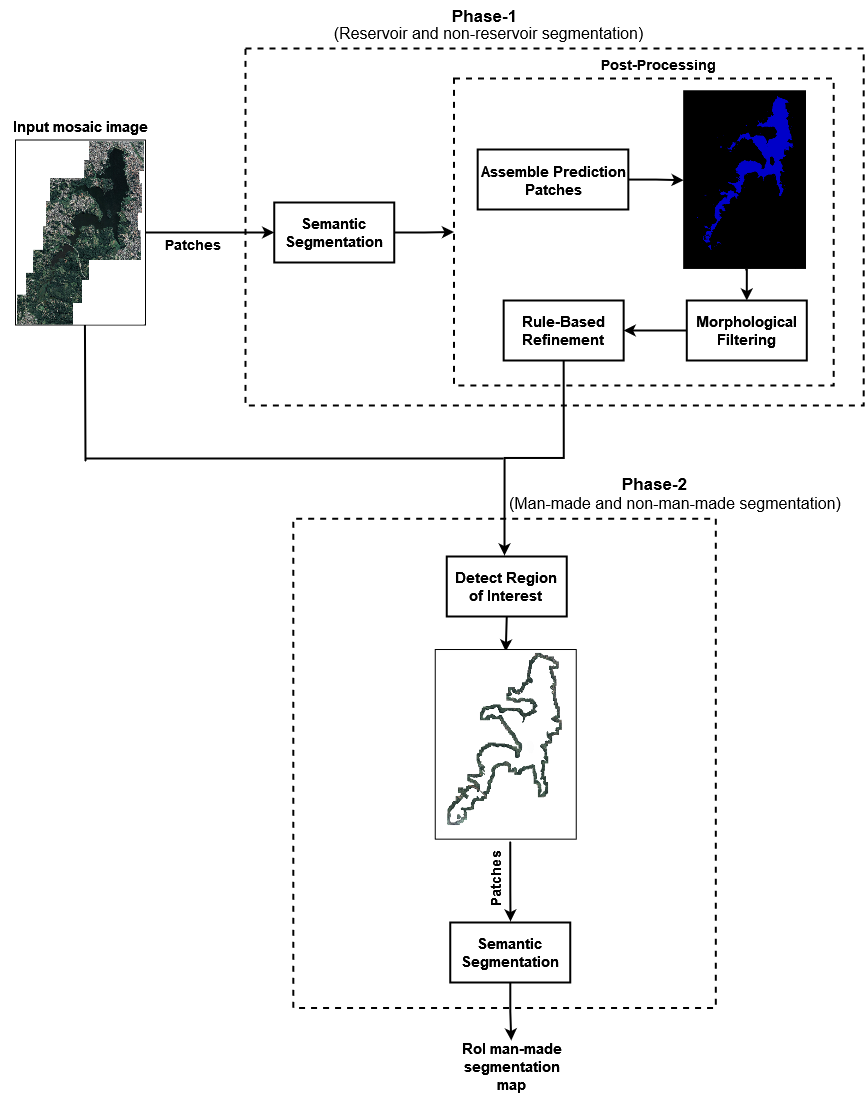}
      \caption{Overview of the proposed analysis workflow.}
      \label{fig-pipeline}
\end{figure}

Our task of interest is man-made object segmentation around reservoirs. The proposed approach (see Figure~\ref{fig-pipeline}) is based on three main steps: 1- Reservoir segmentation; 2- RoIaR detection; 3- Man-made object segmentation in the RoIaR. 
The data is initially collected and pre-processed to be prepared in a suitable manner. The input images are fed to phase-1 for reservoir segmentation. The reservoir map is passed to phase-2, where the RoIaR is detected. This RoIaR is used as a mask where man-made objects are finally segmented. The proposed workflow and implemented steps for preparing that, are detailed in the following subsections.

\subsection{Data Collection}
\label{seccollect}

Our experiments are performed on RGB remote sensing images collected from eight reservoirs in Brazil using the Google Earth Pro\textsuperscript{\textcopyright} software. GE images represent an integration of multiple satellite data sources, mainly DigitalGlobe’s QuickBird commercial satellite and EarthSat, that mostly are from Landsat-7 \cite{qian2020tcdnet}. Original GE imagery has more than three bands. Aiming at improving the appearance of the images, their spectral information is reduced to RGB \cite{c44,c19,c45}. Furthermore, the appearance of GE images is improved using color balancing, warping, and mosaic processing \cite{c11}. Besides being an open dataset of RS images, including historical images and flexibility in selecting images of different resolutions are additional advantages of this platform.

\begin{figure}[!ht]
	\centering 
	\includegraphics[width=0.8\linewidth]{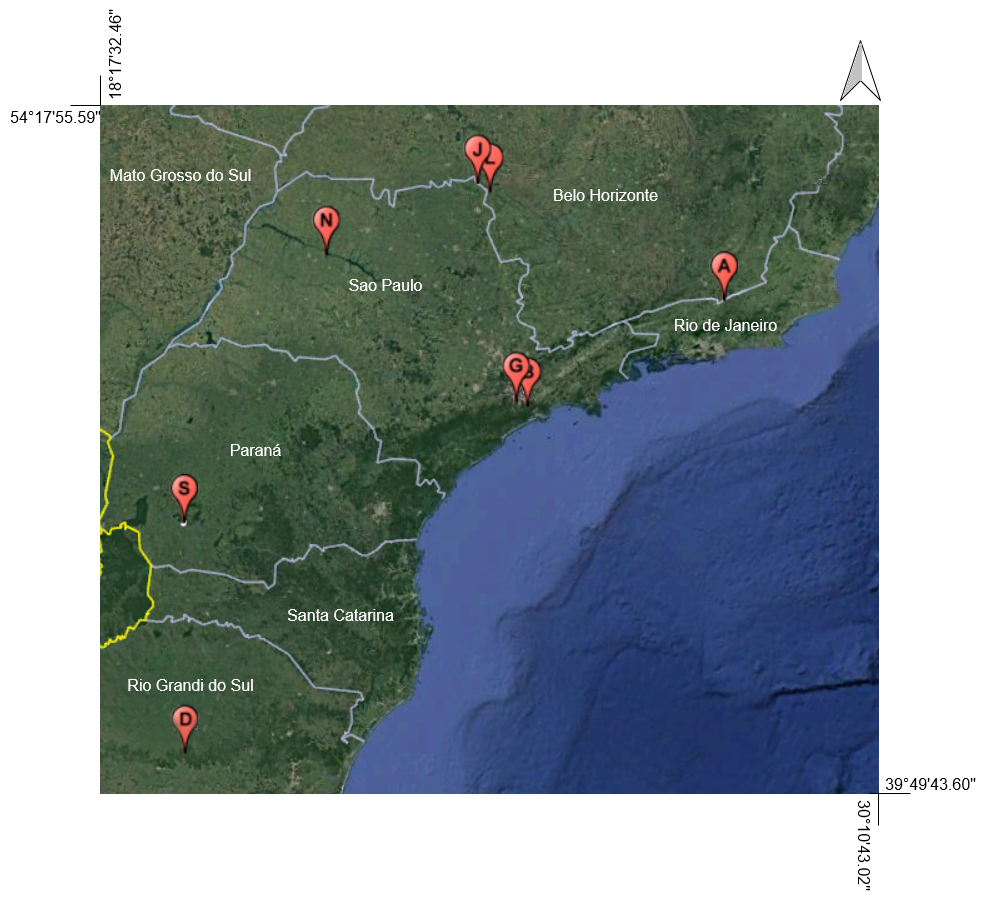}
	\caption{\label{studied areas}Visualization of the studied reservoirs locations.}
\end{figure}

The eight studied reservoirs are Anta, Billings (the largest reservoir in São Paulo, Brazil), Dona Francisca, Guarapiranga, Jaguara, Luiz Barreto, Nova Avanhandav (Nova), and Salto Osório. Their geographic coordinates are listed in Table \ref{tablereservoirs}. Their locations are visualized in Figure \ref{studied areas}. For each reservoir, images over two different years are collected (Table \ref{reseryears}). Totally 206 images with 2683 x 4800 pixel sizes are captured. They are captured in different view altitudes and have consequently different resolutions (from approximately one meter up to two meters).

\begin{table}[!ht]
\centering
\begin{tabular}{c c c}
\hline
\textbf{Reservoir} & \textbf{State} & \textbf{Coordinates}\\
\hline
Anta &   Minas Gerais and Rio de Janeiro & $22^\circ 02'33.20''$ S, $43^\circ 01'16.85''$ W \\
Billings & São Paulo & $23^\circ 48'50.62''$ S, $46^\circ 32'19.39''$ W \\
Dona Francisca  & Rio Grande do Sul & $29^\circ 26'34.18''$ S, $53^\circ 16'09.09''$ W \\
Guarapiranga &  São Paulo & $23^\circ 43'16.93''$ S, $46^\circ 44'22.23''$ W \\
Jaguara &  Minas Gerais and São Paulo & $20^\circ 05'01.85''$ S, $47^\circ 24'10.44''$ W\\  
Luiz Barreto & São Paulo & $20^\circ 14'18.50''$ S, $47^\circ 11'01.95''$ W \\	
Nova & São Paulo & $21^\circ 10'34.54''$ S, $50^\circ 07'34.03''$ W \\	
Salto Osório   & Paraná & $25^\circ 33'28.60''$ S, $52^\circ 57'07.61''$ W\\		
\hline
\end{tabular}
\caption{Locations of the studied reservoirs.\label{tablereservoirs}}
\end{table}

\begin{table}[!ht]
\centering
\begin{tabular}{c c c}
\hline
 &\multicolumn{2}{c}{\textbf{Acquisition Years}}  \\ \cline{2-3} 
\textbf {Reservoir}  & \textbf{Older} & \textbf{Earlier}  \\
\hline
Anta & 2014 & 2020  \\ 
Billings  &  2009 & 2019  \\ 
Dona Francisca  & 2011 & 2017 \\
Guarapiranga  & 2009 & 2019 \\ 
Jaguara &   2010 & 2020\\
Luiz Barreto &  2010 & 2020 \\ 
Nova &  2011 & 2021 \\ 
Salto Osório & 2005 & 2019 \\ 
\hline
\end{tabular}
\caption{\label{reseryears} Acquisition years of each reservoir dataset. Some of the older year images of Luiz and Nova belong to 2004 and 2010, respectively.}
\end{table}

\subsection{Data Preparation and Annotation} 
\label{sec-preprodata}

\begin{figure}[!ht]
\centering
\includegraphics[scale=0.22]{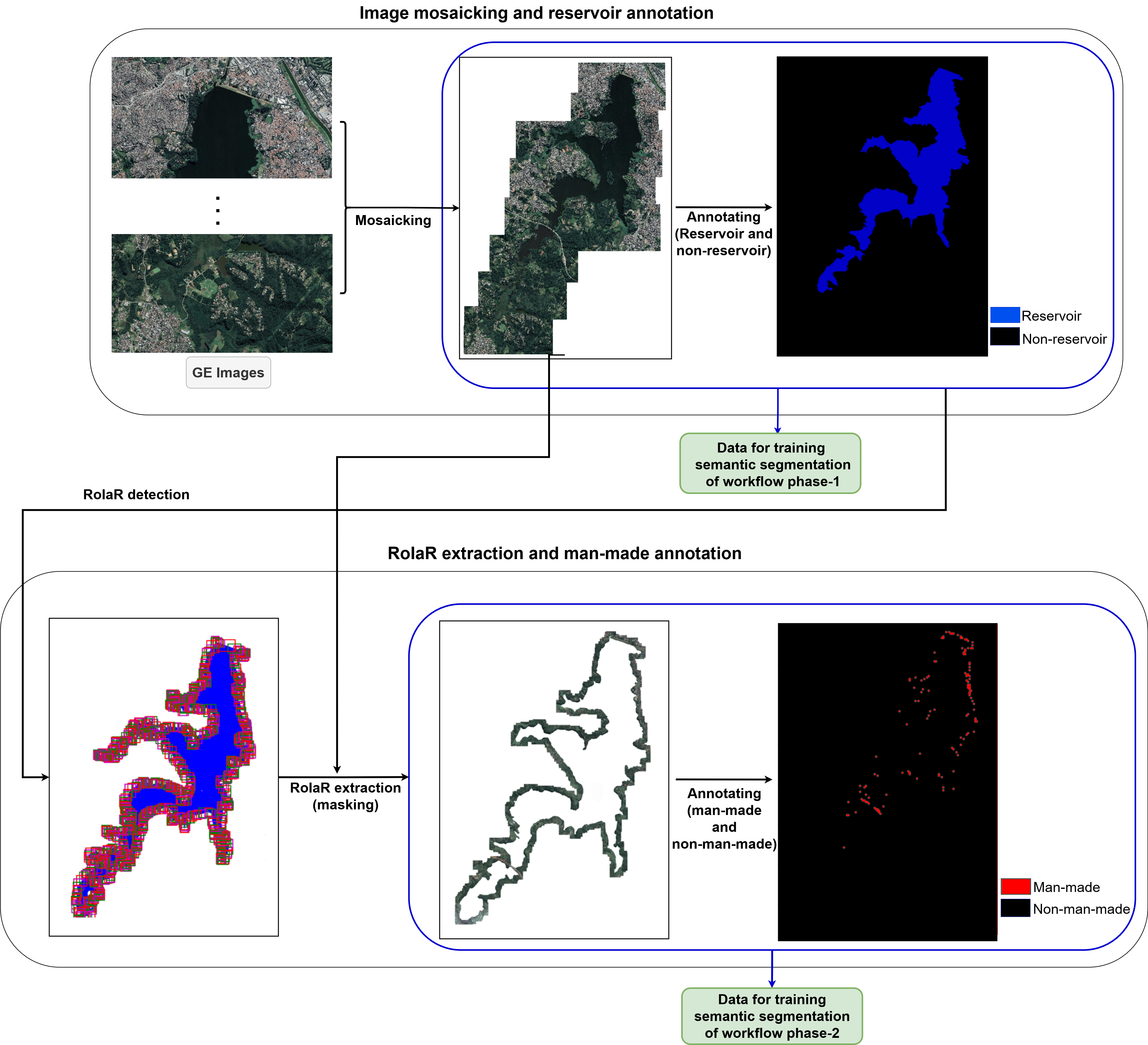} 
\caption{Proposed data preparation and annotation pipeline.}
\label{fig-dataPreparation}
\end{figure}

Data preparation involves two aspects: pre-processing for mosaic image formation and data annotation. The data preparation scheme is illustrated in Figure \ref{fig-dataPreparation} using Guarapiranga reservoir samples. The data preparation aims to prepare data for training the phase-1 and phase-2 semantic segmentation models in Figure \ref{fig-pipeline}. 

As is shown in Figure~\ref{fig-dataPreparation}, the input images are initially mosaicked to eliminate overlapping areas in collected GE images. Constructing the mosaic images is also essential for implementing the next steps. The mosaicked images are annotated into two classes, reservoir and non-reservoir. 

Next, in order to simplify the contour around the reservoirs, a polygonal approximation is initially carried out \cite{polyapproxi,find_contours,rbook}. This allows controlling the coarseness by the polygonal approximation parameter. Then, a rectangular box connecting each pair of consecutive polygon corners is defined. These boxes are enlarged to cover an at least distance from the border of the reservoir. The RoIaR is defined as the union of these boxes (see Figure~\ref{fig-dataPreparation}) and is used to mask the mosaic image.  

The masked RoIaR image is annotated to man-made and non-man-made objects:

\begin{itemize}
    \item {\bf Man-made objects}: road (asphalted and not-asphalted), rooftop, bridge, pool, urban and countryside constructions, impervious surface.
   
    \item {\bf Non-man-made objects}: Vegetation, water body, bare land, plantations, etc.
\end{itemize}

\subsection{Phase-1: Reservoir segmentation}
\label{sec-phase1} 

This step explores a deep neural network that segments input RGB patches to reservoir and non-reservoir. Encoder-decoder-based models have been trained and compared for this step. Below, we briefly describe the two models assessed in this study: U-Net and SegNet. Based on our evaluation, the SegNet-based model has been selected as the best in our experiments.

\begin{figure}[!ht]
      \centering
      \includegraphics[scale=0.37]{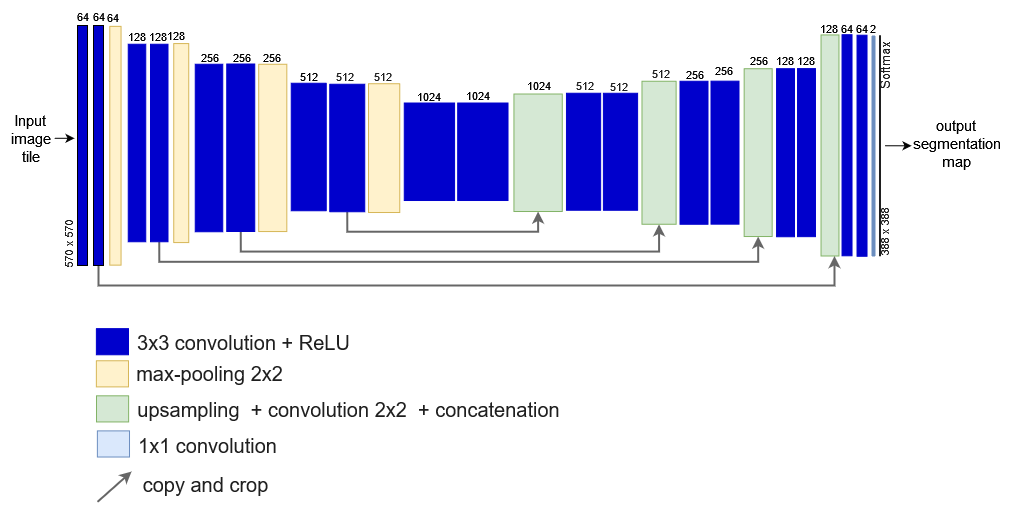}
      \caption{The U-Net architecture.}
      \label{unet}
\end{figure}

The U-Net architecture introduced by~\cite{ronneberger2015u} is based on a downsampling-upsampling procedure that concatenates feature maps between each encoder and corresponding decoder by skip connections (see Figure \ref{unet}). In each step in the encoder path, two 3x3 convolutions followed by a ReLU and a 2x2 max-pooling with stride two are repeated. Furthermore, the number of feature channels in each downsampling step is doubled. After each upsampling in the decoder path, a 2x2 convolution that halves the number of feature channels is applied. These features are concatenated with the cropped feature of the corresponding encoder step, and then two 3x3 convolution-ReLU blocks are implemented.

Due to the unpadded convolutions utilized in the U-Net, the output size of the model is smaller than the input. Therefore, we avoided unpadded convolutions to keep the size of each output equal to the corresponding input (named U-Net\_p). On the other hand, a common strategy in DL research for training the CNNs properly and avoiding training from scratch is utilizing a pre-trained CNN as the initializer or as the fixed feature extractor, called transfer learning. Therefore, we trained a U-Net model whose encoder path was replaced by VGG-16 (named U-Net\_v) and initialized with weights trained on the ImageNet dataset. However, the model overfitted the train set. The last trained U-Net-based model (named U-Net\_s), has fewer features. In this model, there is only one convolution block in each layer that is also batch normalized \cite{batchnorm}.

\begin{figure}[h!]
      \centering
      \includegraphics[scale=0.5]{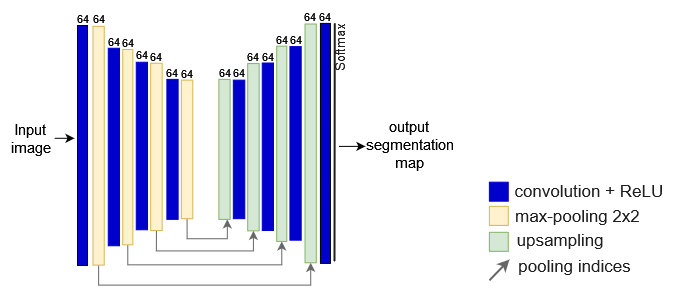}
      \caption{\label{segnet}The SegNet architecture. }
\end{figure}

The SegNet architecture was first introduced by \cite{segbasic}. Similar to the U-Net, SegNet includes an encoder and a decoder part with the advantage that the need for learning to up-sample is eliminated. Since each decoder uses pooling indices computed in the max-pooling step of the corresponding encoder. After each convolution layer in the encoder path, a ReLU not-linearity is used, whereas, in the decoder, no ReLU not-linearity is presented. Furthermore, the number of channels per layer is constant (see figure \ref{segnet}). In the employed architecture (called SegNet\_d), despite the original form, the number of feature channels is doubled at each down-sampling step. Moreover, batch normalization is applied after each convolution layer.

\textbf{Data splitting} Since the collected images correspond to different reservoirs geographically spread in Brazil, they have different visual properties. They may be obtained in different seasons, atmospheric and geological conditions, Etc. A possible approach to address such variability is to adopt the domain adaptation techniques. Since this is out of the scope of this paper, we explored a data splitting approach to ensure variability in the train, test, and validation sets. Therefore, samples from all reservoirs are used in these sets in the following proportions: 60$\%$ for the train set, 20$\%$ for the validation set, and 20$\%$ for the test set.

\textbf{Post-processing}
Feeding models by mosaic images instead of patches is impossible because of the available GPU memory limits. Whereas, in many cases, patches do not contain important information about objects, such as their shapes, sizes, and locations in the images. However, this information is essential for detecting some water bodies from reservoirs. On the other hand, spectral similarities between objects of different classes also cause errors. Therefore, we proposed a post-processing stage to fix these errors.

In this stage, the segmented patches are initially assembled to form the reservoir map. Then, the morphological opening is applied to remove small false positive pixels and other non-interesting water objects such as rivers. Next, morphological closing is applied to remove small false negatives objects inside the reservoir objects.

Applying morphological transformations with a large kernel causes changes in the shapes of objects predicted as the reservoir. Accordingly, in order to remove noisy objects (such as large water bodies around reservoirs) and errors inside reservoirs (such as floating vegetation), two object-based refinements are proposed:

\begin{itemize}
    \item If a non-reservoir object is surrounded by a reservoir object, it is classified as the reservoir.
    \item If the size of a reservoir object is smaller than one-fifth of the size of the largest reservoir object, or the minimum distance between these two objects is greater than 300 meters, then it is classified as non-reservoir.
\end{itemize}

\subsection{Phase-2: Man-made objects segmentation}\label{p2}

Once the reservoir is segmented, the next step is to detect and extract the RoIaR. Two possible approaches for RoIaR detection have been considered: polygonal approximation-based and mathematical morphology-based. The polygonal approximation approach has been described in Section~\ref{sec-preprodata}, which is the one adopted for dataset annotation. Although this approach is useful for sparse data annotation (because we may control the polygonal approximation parameters), it produces patches of varying sizes that may not be suitable for analyzing man-made objects' evolution, for instance. 

Therefore, a mathematical morphological approach is also explored. Let $I$ denote the segmented reservoir image and $s$ a structuring element. The dilated reservoir image is defined as $I_d = I \oplus s$, where $\oplus$ is the morphological dilation. The RoIaR $R$ is defined as $R = I_d - I$, where $-$ denotes set difference. 

Following the data annotation procedure illustrated in Figure~\ref{fig-dataPreparation}, the detected RoIaR is applied as a mask to the original data for RoIaR extraction. The extracted RoIaR is then segmented into man-made and non-man-made objects.  

Two widely used network architectures for RS semantic segmentation are the pyramid networks, and encoder-decoder networks \cite{mou2018rifcn}. In phase-2, the following networks have been assessed: U-Net, Pyramid scene parsing network, Feature pyramid network, and LinkNet, which are detailed in the following.

\begin{figure}[!ht]
      \centering
      \includegraphics[scale=0.46]{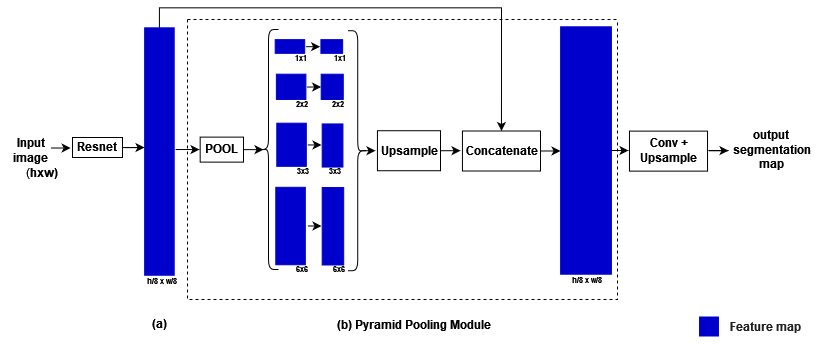}
      \caption{An overview of PSPNet. The size of feature map channels is denoted below each box. The size of the last feature map in (a) is 1/8 of the input image size. }
      \label{psp}
\end{figure}

The Pyramid scene parsing network (PSPNet) has been introduced by \cite{zhao2017pyramid} and won the ImageNet Scene Parsing Challenge 2016. It is a pyramid pooling module that enables the network to capture the context of the whole image. In this module, the feature map is pooled at different sizes and passed through a convolution layer. Next, these features are upsampled and concatenated with the original feature map and passed through a convolution layer to produce the final prediction (see Figure \ref{psp}). We implemented PSPNet with different backbones in this study. Furthermore, besides PSPNet that downsamples input image to 1/8, the 1/4 downsampling is also trained.

\begin{figure}[h!]
      \centering
      \includegraphics[scale=0.3]{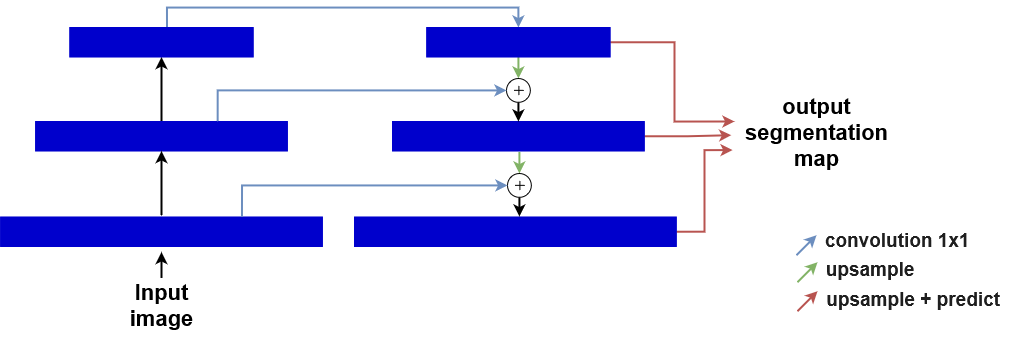}
      \caption{An overview of FPN.}
      \label{fpnf}
\end{figure}

The Feature Pyramid Network (FPN) was initially proposed by \cite{fpnp} for object detection. The general scheme of FPN is illustrated in Figure \ref{fpnf}. The construction of this architecture involves a bottom-up path, a top-down path, and lateral connections. The scaling step in the bottom-up path (and consequently in the top-down path) is two. Each lateral link combines feature maps from the bottom-up and top-down pathways with the same spatial size. Finally, the feature maps in the top-down stages are upsampled to be the same size as the input image. These feature maps are combined and used to produce the prediction map. The ResNet is used as the backbone, whereas in this study, other backbones have also been experimented.

\begin{figure}[h!]
      \centering
      \includegraphics[scale=0.47]{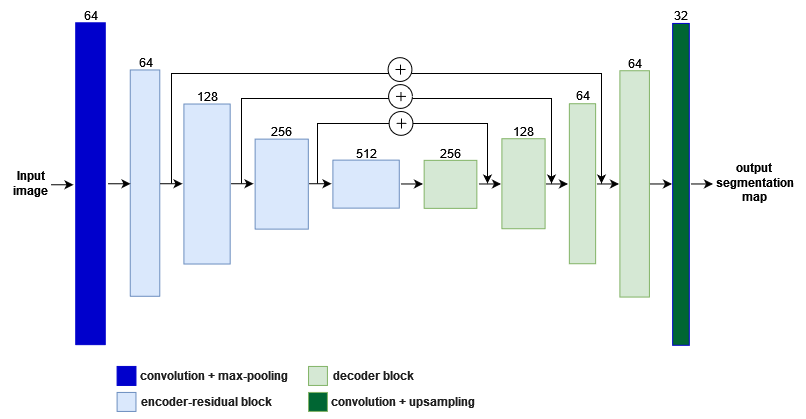}
      \caption{An overview of LinkNet architecture. }
      \label{linknetf}
\end{figure}

The LinkNet architecture proposed by \cite{linknet} is a fast semantic segmentation method that is constructed from an encoder and a decoder path (see Figure \ref{linknetf}). Each residual block in the encoder path consists of two consequent convolution blocks. The input of each residual block is bypassed to its output. The decoder blocks consist of three convolution layers, and the middle is a full convolution. The advantage of the proposed architecture is passing the input of each encoder block to the output of the corresponding decoder block.

\textbf{Data splitting} Splitting data into the train and test sets is reported to work well when the dataset size is modest. On the other hand, the train and test sets must represent possible distributions of the addressed problem. Therefore, 70$\%$ for the train and 30$\%$ for the test set are selected randomly from each RoIaR.

\begin{equation}
    CE(p,y) = 
\begin{cases}
    -\log{p},& \text{if } y = 1\\
    - log(p-1),              & \text{otherwise}
\end{cases}
\label{eqce}
\end{equation}

\noindent where p is predicted probability for class with label y=1. Now lets define a new notation $p_{t}$:

\begin{equation}
    p_{t} = 
\begin{cases}
    p & if y = 1\\
    p-1,              & \text{otherwise}
\end{cases}
\end{equation}

Using this notation we can rewrite Equation \ref{eqce} as $CE(p_{t}) = -log(p_{t})$. To balance the importance of positive/negative examples, we can consider $\alpha_{t}$ as the weight for class 1 and $1-\alpha_{t}$ for class 0, then $\alpha$-balanced CE will be written as:

\begin{equation}
    CE(p_{t}) = -\alpha_{t} log(p_{t})
\end{equation}Finally, to down-weight easy examples, they add factor \texorpdfstring{$(1-p_{t})^\gamma$}{Lg} to CE where \texorpdfstring{$\gamma$}{Lg} > 0 is a tunable parameter. Based on the experiment, \texorpdfstring{$\gamma = 2$}{Lg} works best and is used in this study too.

The dice Loss is based on the dice coefficient (DC); see Equation \ref{dicecoe}. In the case of binary classification, A is the set of all positive examples, and B is the set of correct predicted positive examples. 

\begin{equation}
    DC=\frac {2\left |{ A\bigcap B }\right |}{\left |{ A }\right | + \left |{ B }\right |} 
\label{dicecoe}
\end{equation}

Then, DC can be expressed as the following form:

\begin{equation}
    DC= 2\cdot \frac { TP}{ TP+FP+FN}
\label{dicecoe2}
\end{equation}

\noindent where TP, FP, and FN are true positive, false positive, and false negative, respectively. The dice loss (DL) takes the following form:

\begin{equation} DL=1- 2\cdot \frac{\sum\nolimits_{i=1}^{N}p_{i}r_{i}}{\sum\nolimits_{i=1}^{N}r_{i}+p_{i}}
\label{diceloss}
\end{equation}

\noindent where $p_{i}$ is the predicted probability for pixel i-th and $r_{i}$ is the ground truth of the corresponding pixel. The imbalance between the foreground and background can be efficiently reduced using Dice Loss. However, it disregards the imbalance in data difficulty.

\section{Experimental Results}\label{secresult}

This section describes the experimental evaluation of the proposed workflow. Phases 1 and 2 have been evaluated, and the results are discussed below.

\subsection{Performance Evaluation Metrics}

Three common statistics, precision (Equation \ref{pre}), recall (Equation \ref{rec}), and F1-score (Equation \ref{f1}), are adopted, as well as the confusion matrix of land cover maps. 

\begin{equation} \label{pre}
Precision = \dfrac{TP}{TP + FP} 
\end{equation}

\begin{equation} \label{rec}
Recall = \dfrac{TP}{TP + FN} 
\end{equation}

\begin{equation} \label{f1}
F_{1} = 2 \cdot \dfrac{Precision \cdot Recall}{Precision + recall} 
\end{equation}
\subsection{Phase-1 Experimental Results}

The trained architectures for this phase are modified versions of U-Net and SegNet. All trained models apply the Binary Cross Entropy as the loss function. The learning rate in the Adam optimizer (proposed by \cite{adamref}), is set to 0.001, which is reduced by a factor of 0.2 after every five epochs with no reduction in validation loss down to $10^{-7}$. Although the number of epochs is set to 100, training is stopped after 20 epochs with no reduction in the validation loss. Patches with 416 x 608 pixel sizes are fed into the networks, and train, validation, and test sets contain 6017, 2009, and 1998 patches, respectively. Vertical and horizontal flips are two types of augmentation that each one is applied randomly on 50$\%$ of train patches. The F1-score of trained models in segmenting the train and validation sets are presented in Table \ref{p1models}. As is illustrated in this Table, the U-Net\_v overfits the train set. 

\begin{table}[!ht]
\centering
\begin{tabular}{c c c}
\hline
 & \multicolumn{2}{c}{\textbf{F1-score}} \\ \cline{2-3}
\textbf{Model}  & \textbf{Train set} & \textbf{Validation set} \\ 
\hline
U-Net\_p  & 96.19 & 95.46 \\
U-Net\_v  & 92.11 & 68.86\\ 
U-Net\_s  & 98.16 & 97.80   \\ 
SegNet\_d  & 98.40 & 98.03 \\
\hline
\end{tabular}
\caption{\label{p1models} Performance of trained architectures for phase-1 semantic segmentation stage.}
\end{table}

\begin{figure}[!ht]
       \centering
      \includegraphics[scale=0.95]{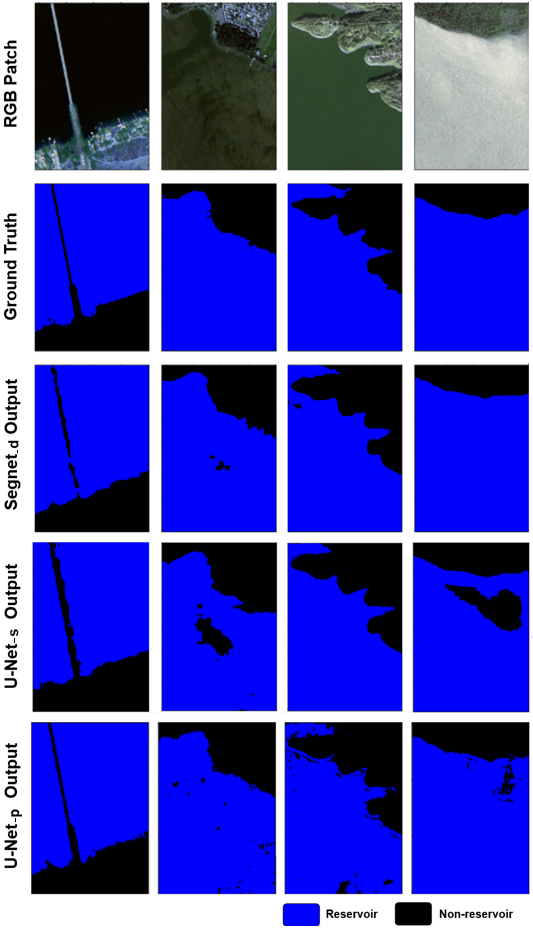}
      \caption{ \label{segsample} Examples of the test set patches beside their corresponding ground truths and segmentation outputs. }
\end{figure}

The performances of models with healthy learning curves in segmenting the validation set are illustrated in Table \ref{p1_best}. As illustrated in the Table, SegNet\_d outperforms the U-Net-based models. The performance of SegNet\_d in segmenting the test set is illustrated in Table \ref{segtestper}. Some patches of studied reservoirs with different spectral properties besides their ground truths and SegNet\_d, U-Net\_s, and U-Net\_p prediction outputs are shown in Figure \ref{segsample}.

\begin{table}[h!]
\centering
\begin{tabular}{c c c c c c c}
\hline
 & \multicolumn{2}{c}{\textbf{Precision}} & \multicolumn{2}{c}{\textbf{Recall}} & \multicolumn{2}{c}{\textbf{F1-score}}  \\ \cline{2-7} 
\textbf{Model} &  \textbf{non-reservoir} & \textbf{reservoir} & \textbf{non-reservoir} & \textbf{reservoir}  & \textbf{non-reservoir} & \textbf{reservoir} \\
\hline
U-Net\_p  &  98.18 & 93.72 & 98.71 &91.27 & 98.44 & 92.48\\
U-Net\_s  &  98.63 & 93.85 & 98.71 & 93.49 & 98.67 & 93.67\\ 
SegNet\_d & 98.79 & 94.39 & 98.82 &94.24  & 98.81 & 94.32  \\
\hline
\end{tabular}
\caption{\label{p1_best} The performance of models with healthy learning curves for Phase-1 semantic segmentation on the validation set.}
\end{table}

\begin{table}[h!]
\begin{center}
\begin{tabular}{c c c c c}
\hline
\textbf{Class} & \textbf{Precision} &\textbf{Recall} & \textbf{F1-score} & \textbf{Support (N.pixels)}  \\
\hline
Non-reservoir & 98.82 & 98.87 & 98.85 & 4211777759  \\ 
Reservoir  &  94.33 & 94.11 & 94.22 &84172385  \\ 
\hline
\end{tabular}
\caption{\label{segtestper} SegNet\_d performance on the test set.}
\end{center}
\end{table}

Besides errors that occur because of spectral similarities between reservoirs and some other objects (such as shadows), there are small water bodies, rivers, Etc., in the images that are segmented as the reservoir by the models. This issue is unavoidable because of feeding patches to the models instead of the original images. Therefore, post-processing the network outputs is an essential task. Morphological operations post-processing is highly effective in removing minor errors, as described above. The applied structuring element size for each reservoir equals 100/(spatial resolution). For example, if the spatial resolution of a mosaic image is one meter, the structuring element size is 100x100.

\begin{figure}[!ht]
      \centering
      \includegraphics[scale=0.23]{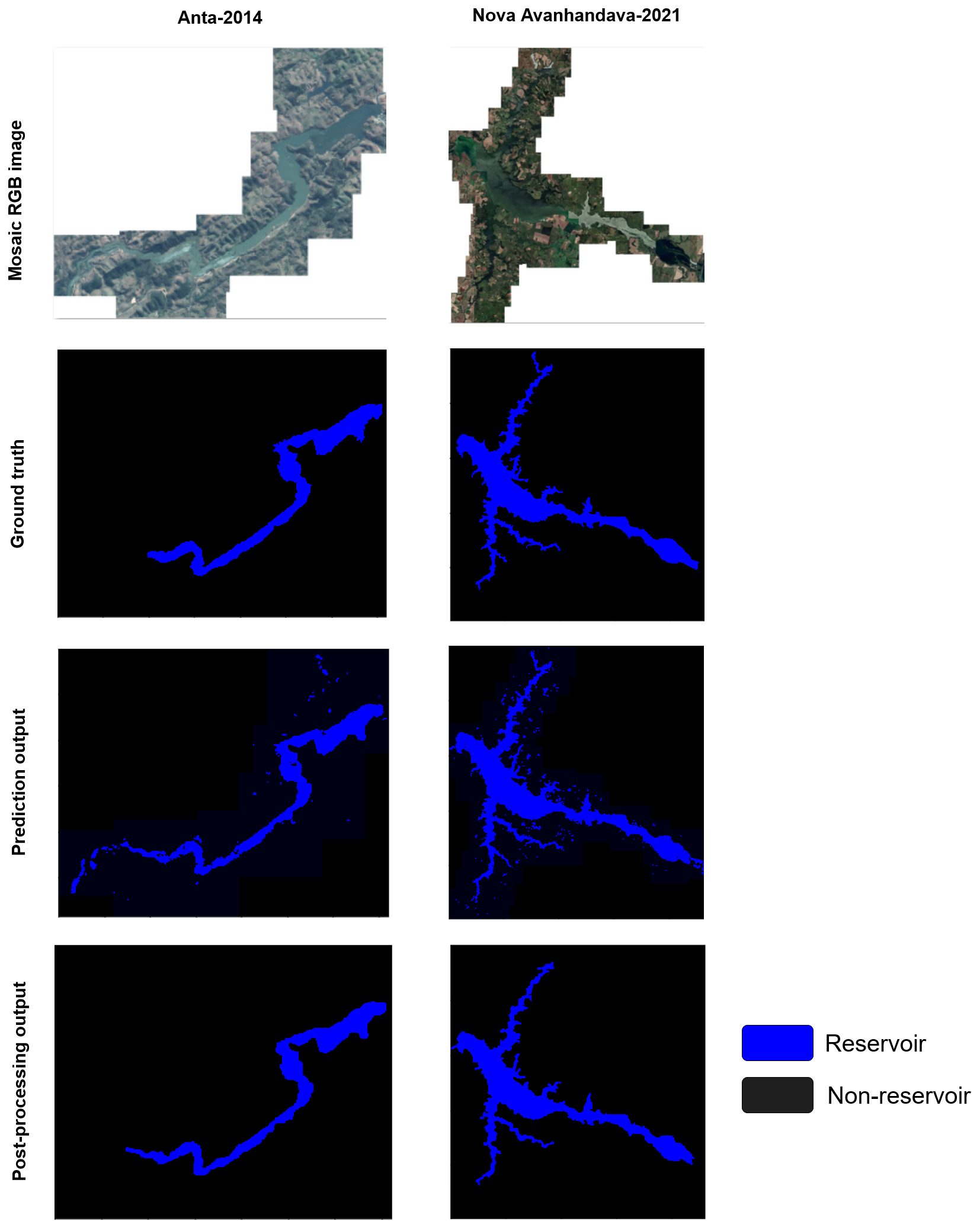}
      \caption{\label{p1examples} Two examples of produced mosaic images, corresponding ground truths, prediction outputs, and post-processing outputs. Anta-2014 with 11687x14430 pixel size, and Nova-2021 with 24830x23193 pixel size are depicted in the first and second columns, respectively.}
\end{figure}

As the reservoirs contain branches, applying morphological operations with large kernel sizes increases FP and FN objects. Accordingly, significant errors are removed by applying the two rules to objects in the produced segmentation maps. Post-processing using only rules is time-consuming because of the high number of FP and FN objects in prediction maps, whereas morphological operations speed up this process. Anta-2014 and Nova-2021 mosaic images, besides their ground truths, model outputs, and post-processing outputs are illustrated in Figure \ref{p1examples}. Moreover, SegNet\_d performance in segmenting these two reservoirs besides post-processing performance are presented in Table \ref{comparper}. Applying the proposed post-processing improves the accuracy of produced reservoir maps except for two of the 16 studied cases.

\begin{table}[h!]
    \centering
        \begin{tabular}{c c c c c c c}
        \hline
       & & \multicolumn{2}{c}{\textbf{Model}}& & \multicolumn{2}{c}{\textbf{Post-processing}} \\\cline{3-4} \cline{6-7}
        
         \textbf{Reservoir} & \textbf{Class}    & \textbf{Precision} & \textbf{Recall}&  & \textbf{Precision} & \textbf{Recall}  \\ 
          \hline
         \multirow{ 2}{*}{Anta-2014}&  non-Reservoir & 98.45  & 98.95& &99.15& 99.56\\
           &Reservoir  & 89.75 & 85.53& &95.72& 92.07 \\
           \hline
           \multirow{ 2}{*}{Nova-2021}& non-Reservoir & 98.33  & 98.03& &98.67& 99.13\\
           &Reservoir  & 92.56 & 93.65 & & 96.63 & 94.90\\
           \hline
        \end{tabular} 
\caption{Prediction and refinement performance metrics for Anta-2014 an Nova-2021.\label{comparper}}
\end{table}

\subsection{Phase-2 Experimental Results}
As discussed above, VGG-16, ResNet-50, and ResNet-101 are the most frequented backbones \cite{9376238}. In this study, these three backbones besides EfficientNetb3 have been experimented. All backbones are initialized with weights trained on the ImageNet dataset. The Adam optimizer is used as the optimizer in all models. The initial learning rate is set to 0.0001 or 0.001, which is automatically reduced by a factor of 0.2 after every five epochs with no reduction in validation loss down to $10^{-7}$. The mini-batch size is set to two and power of two (up to the possible size based on the model's size and available memory). The number of epochs in training all models is set to 80. The vertical and horizontal flips are two augmentation methods that are implemented on different portions of images (up to 0.7). We added dropout regularization (<0.3) to the models with overfitting. Furthermore, experiments are on two train sets, a train set containing 70$\%$ of data or the over-sampled train set. The over-sampled images are images with at least 200 man-made objects pixels.

The evaluation metrics for the highest performance model constructed using each architecture are presented in Table \ref{net2f1}. These models are all trained on the over-sampled train set with a learning rate of 0.0001. Furthermore, the augmentation rate in these models is set to 0.7 for each augmentation method, and the dropout regularization is set to 0.0, 0.3, 0.3, and 0.0, respectively. Regarding the F1-score, the best performance belongs to FPN; however, the differences are insignificant. The utilized backbones for each model in the Table are ResNet50, VGG-16, VGG-16, and Efficientnetb3, respectively. Except for the PSPNet that VGG-16 could improve the performance of the model significantly (2.34$\%$), the performances of the rest models are slightly affected by changing their backbones (<0.73$\%$). In our experiments, oversampling images with more than 200 man-made object pixels improved the performances. Moreover, despite the expectation, increasing batch size did not increase the performance metrics in all cases. Adding the dice loss to the focal loss function significantly improved the models' performances. Although increasing the augmentation rate prevented overfitting in some cases, in other cases increasing dropout and augmentation rates were both essential. Though the FPN outperforms the PSPNet, each epoch training time of PSPNet is less than one-third of the FPN. FPN performance in segmenting test set is presented in Table \ref{fpnper}.

\begin{figure}[!ht]
      \centering
      \includegraphics[scale=0.25]{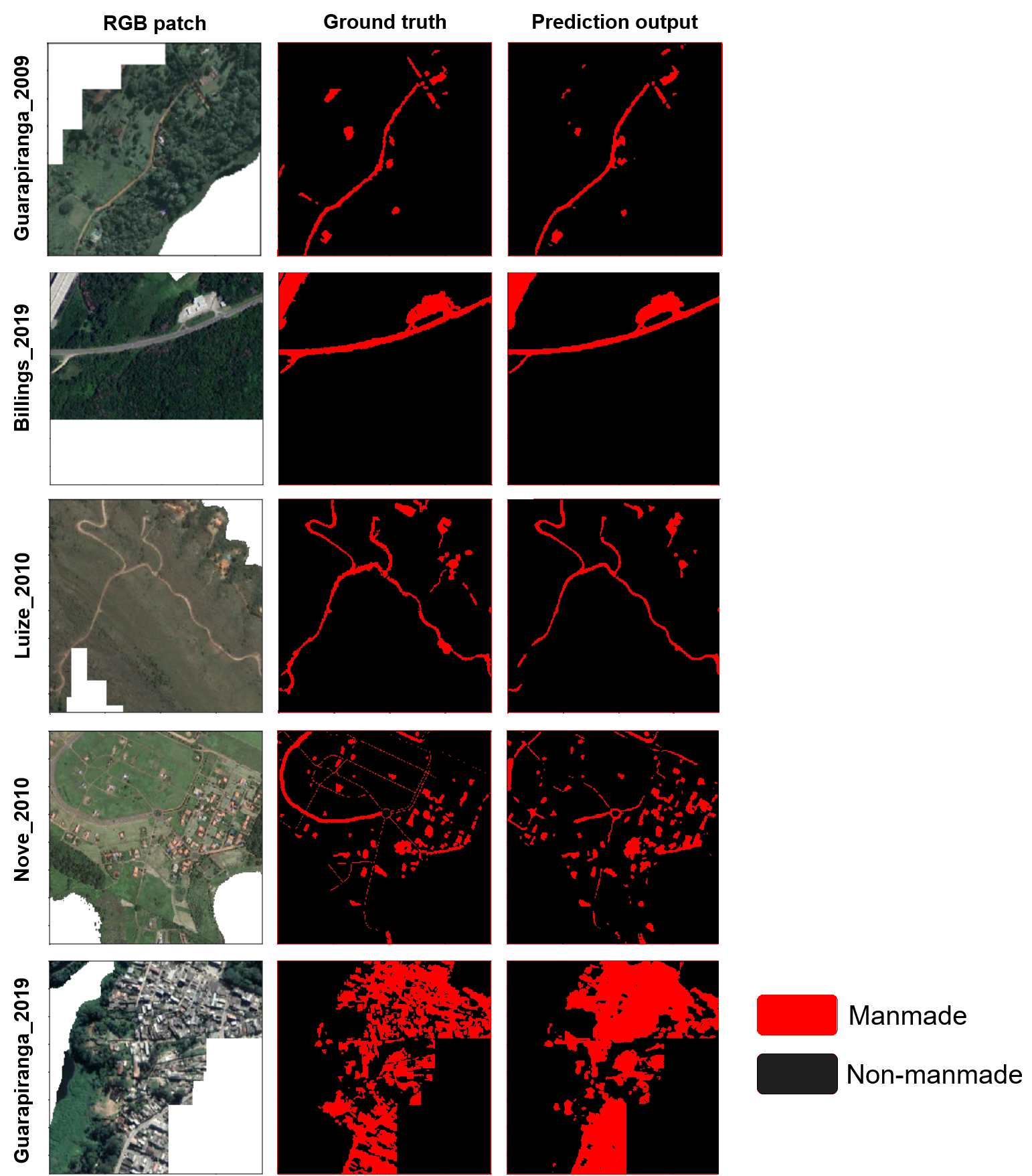}
      \caption{\label{net2wxamples} Samples of four studied reservoirs RoI patches beside their corresponding ground truths and prediction outputs.}
\end{figure}

Moreover, the FPN performance in segmenting RoI of reservoirs located in the countryside and urban areas are computed separately and shown in Table \ref{countryurban}. Some examples of patches besides their ground truths and segmentation outputs are illustrated in Figure \ref{net2wxamples}. This figure illustrates examples of different types of roads, rooftops, and urban and countryside constructions with different density levels. 

\begin{table}[!ht]
\begin{center}
\begin{tabular}{c c c}
\hline
& \multicolumn{2}{c}{\textbf{F1-score}}\\ \cline{2-3}
\textbf{Model} &  \textbf{Train set} & \textbf{Test set}  \\ 
\hline
U-Net  & 91.64  &  90.13  \\ 
PSPNet &  91.29 &  89.58    \\
FPN & 92.16 & 90.32  \\
LinkNet & 91.95 & 90.15  \\
\hline
\end{tabular}
\caption{The highest achieved performances using trained models for Phase-2 semantic segmentation on train and test sets.\label{net2f1}}
\end{center}
\end{table}

\begin{table}[h!]
\begin{center}
\begin{tabular}{c c c c c}
\hline
\textbf{Class} & \textbf{Precision} &\textbf{Recall} & \textbf{F1-score} & \textbf{Support (N.pixels)} \\
\hline
non-man-made & 99.52 & 99.56 & 99.54 & 327065669\\ 
man-made  & 81.79 & 80.43 & 81.10 & 8101819\\ 
\hline
\end{tabular}
\caption{\label{fpnper} FPN performance in segmenting test set into the man-made and non-man-made objects pixels. }
\end{center}
\end{table}

\begin{table}[h!]
\begin{center}
\begin{tabular}{c c c c c c c}
\hline
&\multicolumn{2}{c}{\textbf{Precision}} & \multicolumn{2}{c}{\textbf{Recall}}& \multicolumn{2}{c}{\textbf{F1-score}} \\
\textbf{Class} & C & U & C & U &C&U  \\
\hline
non-man-made & 99.68 & 99.39 & 99.73 & 99.26 &99.71 &99.33\\ 
man-made  & 78.70 & 86.62 & 75.78 & 88.75 &77.21 &87.67\\ 
\hline
\end{tabular}
\caption{\label{countryurban} FPN performance in segmenting countryside and urban man-made objects. C and U are the abbreviations for countryside and urban. }
\end{center}
\end{table}

\subsection{Workflow evaluation}
\label{sec-workEval}

\begin{figure}[!ht]
      \centering
      \includegraphics[scale=0.61]{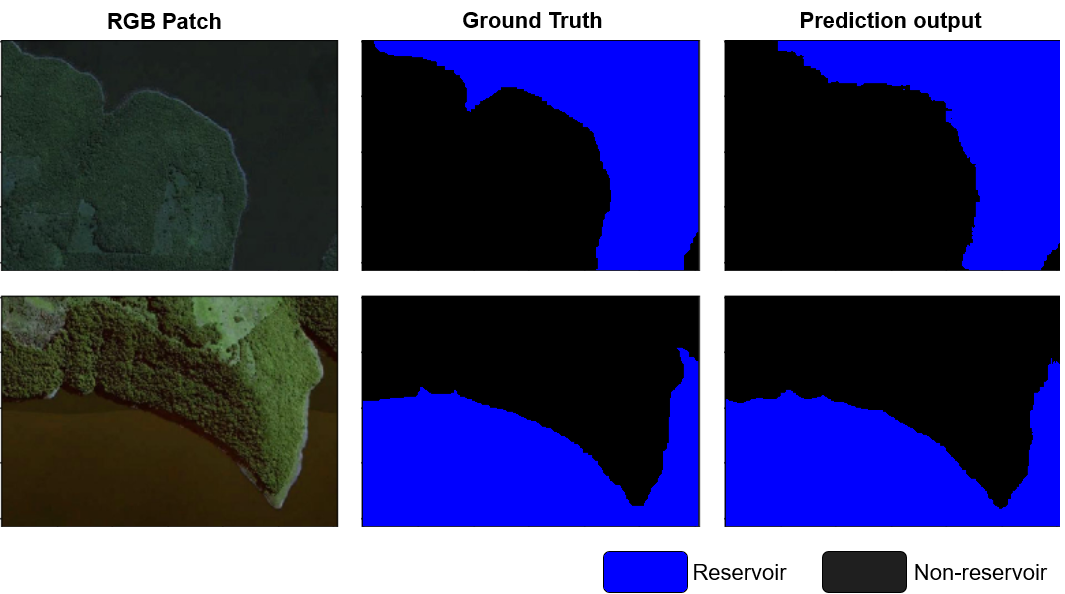}
      \caption{Two samples of Barra phase-1 patches, besides their corresponding ground truths and semantic segmentation results.}
      \label{p1evalsam}
\end{figure}

\begin{figure}[h!]
      \centering
      \includegraphics[scale=0.85]{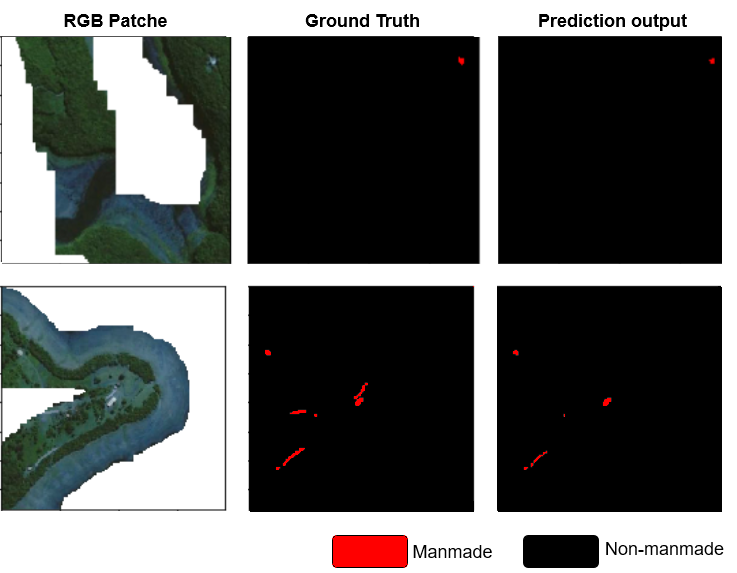}
      \caption{\label{p2evalsam} Two samples of Barra RoI patches, their corresponding ground truths, and semantic segmentation results. }
\end{figure}

We evaluated the proposed workflow using a dataset collected from the Barra Grande reservoir (Barra). Barra is located in Santa Catarina and Rio Grande do Sul states in Brazil. The collected images belong to 2021, and their spatial resolution is two meters. To evaluate the proposed workflow using the collected data, first, patches with 416 x 608 pixel size are constructed from the mosaic RGB image of Barra. Next, patches are fed to the trained SegNet\_d to be segmented into the reservoir and non-reservoir. The SegNet\_d performance is evaluated by comparing model outputs with manually produced ground truths. In the next step, the SegNet\_d outputs are assembled to be refined using the proposed post-processing stage. The refined reservoir map is used to detach the RoI around Barra. The covered distance from the border of the reservoir is 200 meters. In  Table \ref{p1eval}, the performances of the phase-1 semantic segmentation stage, besides the performance of proposed post-processing, are reported. Table \ref{p2eval} shows the evaluation metrics for the phase-2 semantic segmentation stage. Furthermore, some samples of phase-1 and phase-2 semantic segmentation outputs are illustrated in Figures \ref{p1evalsam} and \ref{p2evalsam}, respectively. 

\begin{table}[h!]
    \centering
\begin{tabular}{c c c c c c c c}
    \hline
        & \multicolumn{3}{c}{\textbf{Model}}& & \multicolumn{3}{c}{\textbf{Post-processing}} \\\cline{2-4} \cline{6-8}
        
       \textbf{class}    & \textbf{Precision} & \textbf{Recall} &\textbf{F1-score} & &\textbf{Precision} & \textbf{Recall}&\textbf{F1-score}  \\ 
        \hline
       non-Reservoir & 98.39  & 96.86 &97.62 & &98.38 &98.36&98.37 \\
        Reservoir  & 84.00  & 91.21 &87.45 & & 90.92 &91.04&90.98  \\
           \hline
    \end{tabular}
    \caption{\label{p1eval} Performance of phase-1 semantic segmentation and post-processing stages in segmenting Barra dataset to reservoir and non-reservoir.}
\end{table}
  
\begin{table}[h!]
\begin{center}
\begin{tabular}{c c c c}
\hline
\textbf{Class} & \textbf{Precision} &\textbf{Recall} & \textbf{F1-score}  \\
\hline
non-man-made & 99.99 & 99.99 & 99.99  \\ 
man-made  & 73.29  & 79.43 & 76.23 \\ 
\hline
\end{tabular}
\caption{\label{p2eval} Performance of phase-2 semantic segmentation stage in segmenting Barra RoI to man-made and non-man-made.}
\end{center}
\end{table}

\subsection{Benchmark} 

In order to show the effectiveness of our proposed two-phase approach, we applied a single-phase network for semantic segmentation of reservoir, man-made, and non, as the baseline. In this model, the VGG-16 is used as the backbone, the learning rate is set to 0.0001, the number of epochs is set to 150, the early stopping is not applied, and the summation of Dice and Focal losses is used as the loss function. The learning rate is reduced by a factor of 0.2 after every five epochs with no reduction in validation loss down to $10^{-7}$. Same as the phase-2 training phase, we constructed patches with 384 x 384 pixel size and split them into two sets, train and test. 

Since man-made objects inside RoIaR are annotated as man-made and outside as non-man-made (because they are not around the reservoir), the baseline performance is poor (see Table  \ref{benchmark}), as expected. This simple baseline approach illustrates the importance of our proposed two-phase approach. 
   
\begin{table}[!ht]
\centering
\begin{tabular}{c c c c c c c}
\hline
 & \multicolumn{2}{c}{\textbf{Precision} }& \multicolumn{2}{c}{\textbf{Recall}} & \multicolumn{2}{c}{\textbf{F1-score}}  \\
\textbf{Class} & \textbf{Train} & \textbf{Test} &\textbf{Train} & \textbf{Test} & \textbf{Train} & \textbf{Test} \\
\hline
Reservoir &96.58& 96.15  & 96.52 & 95.72 & 96.55 &95.94  \\ 
Man-made   &62.17& 59.02 &50.78  & 49.13 &  55.90  &53.62 \\ 
Non       &98.64& 98.37 & 98.88& 98.69 & 98.76 &98.53 \\ 
\hline
\end{tabular}
\caption{U-Net performance in segmenting train and test sets into reservoir, man-made, and non. \label{benchmark}}
\end{table}

\section{Discussion}\label{secdiscuss}

The experimental performance evaluation has addressed the results of phases 1 and 2 of the proposed workflow, workflow validation by an external testing dataset, and the single-phase segmentation benchmark result. 


Reservoir segmentation is addressed in phase-1 of the workflow. We trained three U-Net-based models in this phase. The vanilla U-Net was changed to keep the size of each output equal to the corresponding input to produce a pixel-wise classification. Besides, a U-Net with VGG-16 as the backbone was trained. The model over-fitted highly to the train set. Decreasing the number of feature maps in the model (named U-Net\_s) caused performance improvement and fixed the over-fitting issue, as shown in Table \ref{p1models}. A SegNet-based architecture was also trained to examine its ability to enhance segmentation outputs. However, it outperformed the U-Net\_s slightly (1.23$\%$ in F1-Score). 

The reservoirs are considered in a broad class called water bodies. In this study, a post-processing stage is proposed to eliminate errors caused by floating vegetation, and delete FP and FN objects caused by spectral similarities between reservoirs and other objects. The proposed post-processing improved the overall accuracy and provided a clear map of the reservoirs, as shown by the examples in Table \ref{comparper} and Figure \ref{p1examples}.


Phase-2 restricts the segmentation of man-made objects in the RoIaR. Four DL architectures have been evaluated to segment the man-made objects: U-Net, FPN, LinkNet, and PSPNet. This problem typically involves imbalanced data because of government policies to protect such areas besides difficulty in segmenting countryside man-made objects.

In order to address these issues, we tried out the capability of two recommended loss functions (Dice and Focal losses) and the over-sampling strategy. Although Focal loss was reported as the best loss function for segmenting unbalance data, adding Dice loss to the Focal loss significantly improved the performances. Furthermore, oversampling improved the performances as well. We trained each architecture with four different backbones, ResNet50, ResNet101, VGG-16, and EfficientNetb3. The highest improvement caused by changing the backbone belongs to VGG-16 in PSPNet, 2.24$\%$, whereas changing the backbone in other architectures had a low contribution.


Workflow validation has been carried out using data not seen by the model during training (Barra reservoir, see Section~\ref{sec-workEval}). The validation data included realistic noise and difficulties such as clouds. Despite this, the phase-1 model achieved to 92.54$\%$ average F1-score that was even improved to 94.68$\%$ by applying post-processing techniques (see Table \ref{p1eval}). Additionally, the reservoir is in the countryside. The majority of roads are not asphalted, and man-made objects present different visual features from urban areas. Also, there are fewer samples of them in the training data. Accordingly, segmenting them is more complicated compared to urban areas. Nonetheless, the phase-2 model could gain an acceptable performance, as seen in Table \ref{p2eval}.

We increased the feature maps in the phase-2 trained U-Net-based model and trained that to segment collected data into the reservoir, man-made and non. The data was split into the train and test sets, and no early stopping was applied. Nonetheless, the model man-made F1-score was 35.74 $\%$ less than the phase-2 U-Net model.

\section{Conclusions}\label{secconclud}

In this study, we proposed a two-phase workflow to segment man-made objects around reservoirs in an end-to-end procedure. In order to improve produced reservoir maps, a post-processing stage is proposed that, besides increasing the precision metric, its effect is remarkable by visual evaluation. A small portion of images belongs to the class of man-made objects, specially countryside man-made objects. Nonetheless, we gained promising results by collecting images of reservoirs mainly located in the countrysides, and defining a suitable loss function. The collected RS images have high spatial resolutions, contain reservoirs with different spectral properties, contain urban areas as well as countrysides, and are acquired from different states and seasons. These factors increase the reliability and robustness of constructed models and the proposed workflow. The trained workflow was evaluated with an external testing dataset. Although the collected images are noisy in some areas and the RoIaR is in the countryside, the average F1-scores of phase-1 and phase-2 outputs show the reliability of the prepared workflow. The workflow outperformed significantly in man-made objects segmentation compared to the single-phase segmentation benchmark.

We suggest two relevant directions for future research: change detection and domain adaptation. An important possible application of RoIaR man-made objects segmentation is the timely detection of unauthorized constructions around the reservoirs. This social problem might lead to serious consequences such as reservoir contamination and dangerous situations for communities living in such places. Unfortunately, if such constructions are not detected in their first stages and local communities start to live there, it becomes more and more difficult for public services to move such communities. Hence, timely man-made object change detection in the RoIaR is an important application that might rely on the segmentation procedure described in this paper.

On the other hand, a key issue of remote sensing imaging is the challenges in analyzing data from different locations and dates. Geographical and atmospheric variations affect the images, and domain adaptation approaches must often be developed. This problem has been circumvented in this paper by sparse annotation of all considered reservoirs, reflected by our sampling strategy. We are considering other possible domain adaptation approaches such as few-shot and self-supervised learning~\cite{monteiro2022self}. 

\vspace{1cm}
\hspace{-0.7cm}
\textbf{\Large Acknowledgements:}
 The authors would like to thank CAPES and FAPESP $\#$2015/22308-2 for their support during the development of this work.

\bibliographystyle{alpha}
\bibliography{sample}

\end{document}